\documentclass[10pt,twocolumn,letterpaper]{article}

\usepackage{graphicx}
\usepackage{subcaption}
\usepackage{float}
\usepackage[justification=raggedright]{caption}	
\usepackage{lscape}                                         

\usepackage{animate}

\usepackage{booktabs}                   
\usepackage{multirow}

\usepackage{paralist}
\usepackage{enumitem}
\usepackage{wasysym}

\usepackage{bm}                          
\usepackage{epsfig}                      
\usepackage{graphicx}                  
\usepackage{times}
\usepackage{units}
\usepackage{color}

\usepackage{comment}

\usepackage{url}  
\usepackage[pagebackref=true,breaklinks=true,letterpaper=true,colorlinks,bookmarks=false]{hyperref}

\usepackage{xspace}
\usepackage[table]{xcolor}
\usepackage{setspace}

\newlength\paramargin
\newlength\figmargin
\newlength\secmargin
\newlength\figcapmargin

\newcommand{\mpage}[2]
{
\begin{minipage}{#1\linewidth}\centering
#2
\end{minipage}
}

\newcommand{\mfigure}[2]
{
\begin{minipage}{#1\linewidth}\centering
\includegraphics[width=\linewidth]{#2}
\end{minipage}
}

\newcommand{\figref}[1]{Figure~\ref{fig:#1}} 
\newcommand{\tabref}[1]{Table~\ref{tab:#1}}

\long\def\ignorethis#1{}

\newcommand{\tb}[1]{\textbf{#1}}

\def\xi{\mathbf{x}_i}

\graphicspath{{figure}, {example}}

\usepackage{cvpr}

\cvprfinalcopy 

\def\cvprPaperID{****} 

\ifcvprfinal\fi

\begin{document}

\title{Deep Paper Gestalt}

\author{Jia-Bin Huang \\ Virginia Tech \\ \url{jbhuang@vt.edu}}


\maketitle

\begin{abstract}

Recent years have witnessed a significant increase in the number of paper submissions to computer vision conferences.
The sheer volume of paper submissions and the insufficient number of competent reviewers cause a considerable burden for the current peer review system.
In this paper, we learn a classifier to predict whether a paper should be accepted or rejected based solely on the visual appearance of the paper (i.e., the gestalt of a paper).
Experimental results show that our classifier can safely reject 50\% of the bad papers while wrongly reject only 0.4\% of the good papers, and thus dramatically reduce the workload of the reviewers.
We also provide tools for providing suggestions to authors so that they can improve the gestalt of their papers.
\end{abstract}

\section{Introduction}
\label{sec:intro}

\begin{figure}[t]
\includegraphics[width = \linewidth]{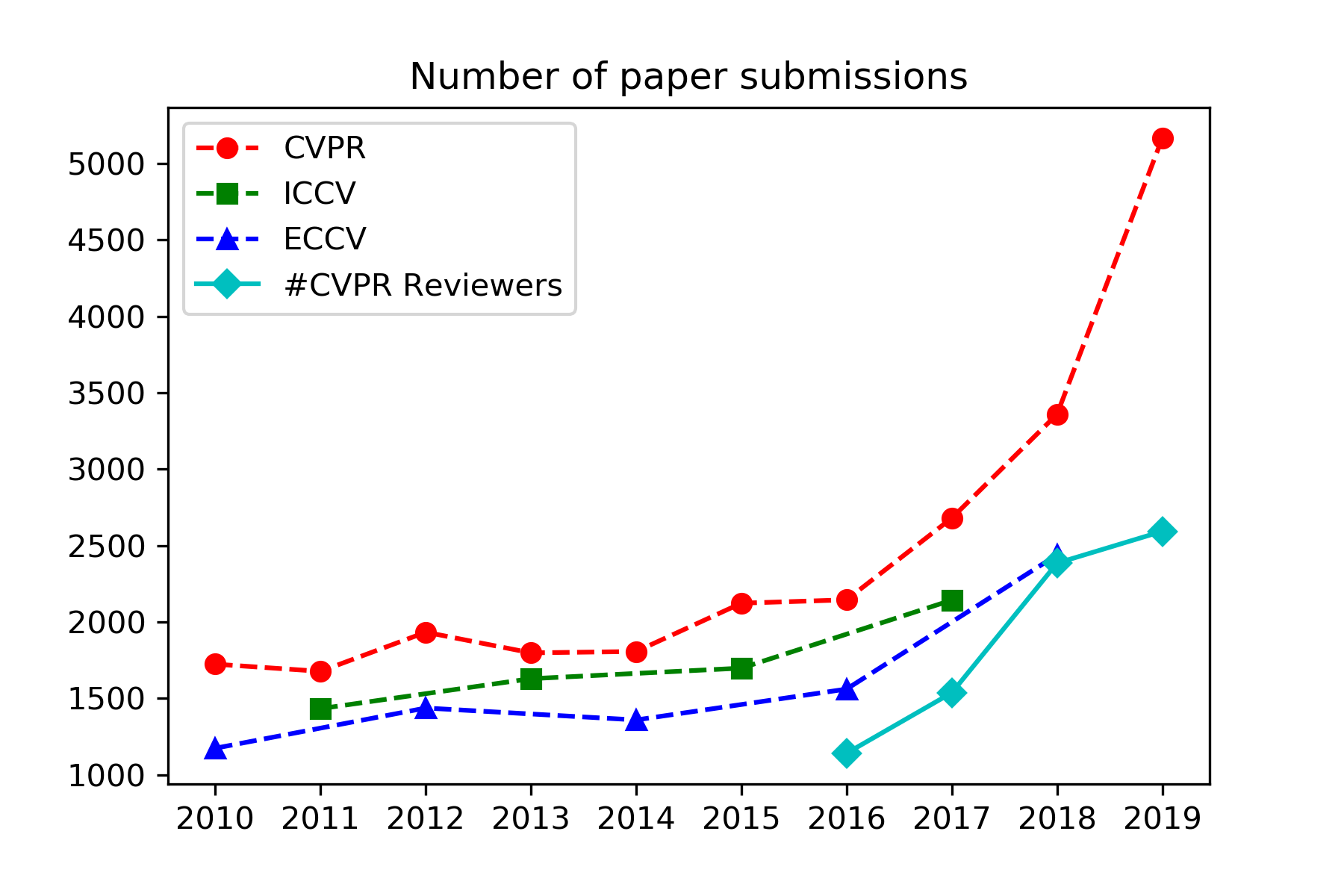}
\vspace{\figcapmargin}
\caption{\tb{Need.} The number of the paper submissions to top-tier computer vision conferences has been increased dramatically over the past few years. The number of competent reviewers (as shown in the cyan curve), however, does not grow nearly as fast.}
\label{fig:motivation}
\end{figure}

Peer review --- a thorough examination of a scholarly work by other experts in the community --- is an essential aspect of disseminating scientific results.
However, the record-breaking number of paper submissions to top-tier computer vision conferences and the insufficient number of competent reviewers make the peer review process increasingly more difficult (see \figref{motivation}).
To review all these submissions, conference organizers have to expand the pool of reviewers and inevitably include less experienced students~\cite{NIPS2018}. 
Consequently, the authors who spent months or years of efforts on a paper submission may end up receiving poorly justified, ill-considered, or unfair reviews.

In this paper, we address this pressing issue in two aspects.
First, we train a deep convolutional neural network using prior conference proceedings to determine the quality of the paper based on its visual appearance (known as paper gestalt~\cite{Bearnensquash2010}).
Second, we provide diagnostic tools to help authors enhance their future paper submissions.
Trained on ICCV/CVPR conference and workshop papers from 2013 - 2017, our deep network based classifier achieves 92\% accuracy on papers in CVPR 2018.
Our model safely rejects the number of bad paper submissions by 50\% while sacrificing only 0.4\% of good paper submissions.
Our system can thus be used as a \emph{pre-filter} in a cascade of the paper review process.
Using the collected Computer Vision Paper Gestalt (CVPG) dataset, we can 1) visualize class-specific discriminative regions of a particular paper submission or 2) translate a bad paper to a good one directly.
These tools help inform the authors \emph{where} and \emph{how} to improve the gestalt of their papers.
%











\section{Related Work}
\label{sec:related}
%

\paragraph{Administrative methods.}
Several methods have been proposed to address the surge in the number of paper submissions through administrative policies.
Examples include desk-reject by area/program chairs (e.g., violation of anonymity, formatting, or clearly out of scope), mandatory abstract submission one week before the paper submission deadline, expansion of the reviewer pool, and training materials for inexperienced reviewers~\cite{CVPRreview}.

\paragraph{Text-based methods.} Automatic grading techniques have been developed for grading essay~\cite{larkey1998automatic}, response to mathematical questions~\cite{lan2015mathematical}, and handwritten work~\cite{singh2017gradescope}.
These techniques, however, do not take into account the rich visual information available in the paper and may be subject to bias toward popular keywords trending in the community.

Our tool for improving paper gestalt is related to sentence editing~\cite{guu2018generating,weston2018retrieve} and automatic random paper generator for computer science~\cite{SCIgen} and math~\cite{Mathgen}.
Our approach differs in that we directly learn the mapping in the image space.

\paragraph{Vision-based methods.} Computer vision techniques have been applied to accessing the quality of actions~\cite{pirsiavash2014assessing}, surgical skills~\cite{zia2016automated}, and images~\cite{wang2004image,zhang2018unreasonable}.
The work most related to our work is that of the awesome Bearnensquash~\cite{Bearnensquash2010}, where the AdaBoost algorithm is used for learning the good/bad paper classifier.
Building upon the methodology in \cite{Bearnensquash2010} that relies on hand-crafted visual features, we revisit the paper gestalt problem with deep learning and learn task-specific representation through an end-to-end training process.
%



 


\section{Learning to Recognize Good/Bad Papers}
\label{sec:classification}

We leverage deep convolutional neural networks (ConvNets) to learn discriminative representation based solely on the visual appearance of a paper, known as \emph{paper gestalt}.
In the following, we start with describing the problem formulation and presenting our dataset construction process.
We then provide the implementation details of the network training.
We validate the performance through an empirical evaluation on a held-out testing set and visualize the class-specific discriminative regions produced by the trained network.

\subsection{Problem formulation}
We formulate the problem as a binary classification task. 
Our training dataset consists of N labeled data samples, $\{(x_1, y_1), (x_2, y_2), \cdots, (x_N, y_N)\}$, where $x_i$ denotes the i-th paper and $y_i \in {0, 1}$ is binary label indicating whether the i$^{th}$ paper $x_i$ is a good paper or a bad one.
Our goal here is to learn a function $F_\theta(\cdot)$ parametrized by $\theta$ that can recognize good/bad papers from unseen paper submissions (e.g., paper submissions to future conferences).

\subsection{Dataset construction}

\paragraph{Data source.} We collect positive examples (good papers) from the list of accepted papers in top-tier computer vision conferences. 
Specifically, we gather the Open Access versions of the accepted papers from recent conferences sponsored by the \href{https://www.thecvf.com/}{Computer Vision Foundation (CVF)}.
This includes six CVPR and three ICCV proceedings from 2013 to 2018.

For negative examples, as we do not have access to papers that were rejected from these conferences, we follow \cite{Bearnensquash2010} and use workshop papers as an approximation.
Similar to the conference papers, we gather the Open Access versions of all the workshop papers from the CVF website.
Note that these negative examples can be noisy as some of the papers 1) were also accepted at the main conferences or 2) were not submitted to the main conference.
At the same time, these workshop papers can also be viewed as ``hard negative'' examples as many of the papers have been significantly improved by addressing the comments from the reviewers.

\paragraph{Data acquisition and preprocessing.} Here we outline the detailed steps we used for constructing the dataset.

\begin{enumerate}
\item \tb{Crawl}: We crawl both the positive and negative examples from the \href{http://openaccess.thecvf.com/menu.py}{CVF Open Access website}. 
\item \tb{Filter}: As some of the workshop papers have different page limits as the main conference papers (e.g., 6 pages including references), the classification task becomes trivial for these cases. We therefore keep only papers with sufficient ($\geq 7$) pages.
\item \tb{PDF2Image}: We use the \href{https://github.com/Belval/pdf2image}{pdf2image}, a python wrapper for \href{https://linux.die.net/man/1/pdftoppm}{pdftoppm}, to convert the downloaded PDFs to images. We arrange these pages into a $2 \times 4$ grid. For papers with a missing 8$^{th}$page, we pad it with a blank page. We also discard the pages greater than 8 (mainly references cited in the paper). The original size of the converted image is of size $2200 \times 3400$ pixels.
\item \tb{Pre-processing}: To prevent the data leakage problem, we remove the header on top of the first page. Without this preprocessing step, the learned classifier can become overly optimistic or even invalid because the classifier can focus on the header region while ignoring the visual contents of the paper.
\end{enumerate}

\begin{figure}[t] \centering
\includegraphics[width =\linewidth]{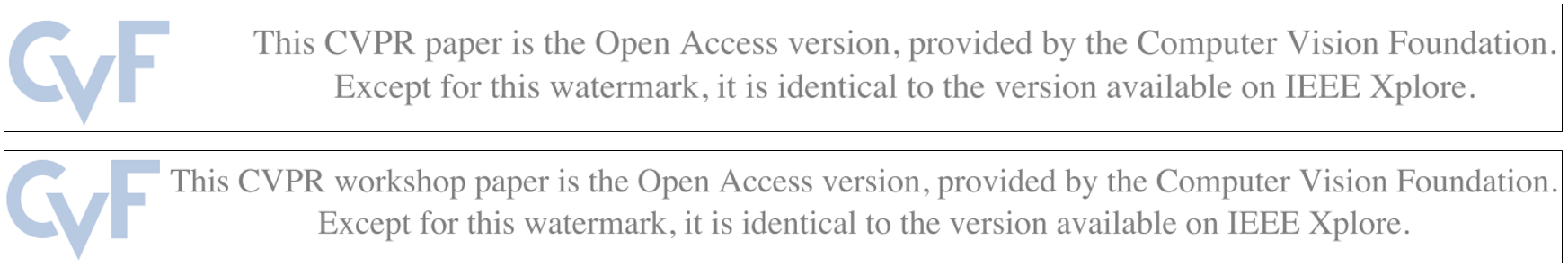} 
\vspace{\figcapmargin}
\caption{\tb{Preventing data leakage.} The header of the Open Access versions of the papers contains the information that the classification models try to predict (i.e., good or bad papers). While the differences are small in the resized images, our results show that our classification network can easily achieve 100\% accuracy on both the training and testing set when the headers were not removed, suggesting that the network found a way to ``cheat''.}
\label{fig:header}
\end{figure}
 \begin{figure*}[t] \centering
\includegraphics[width =\linewidth]{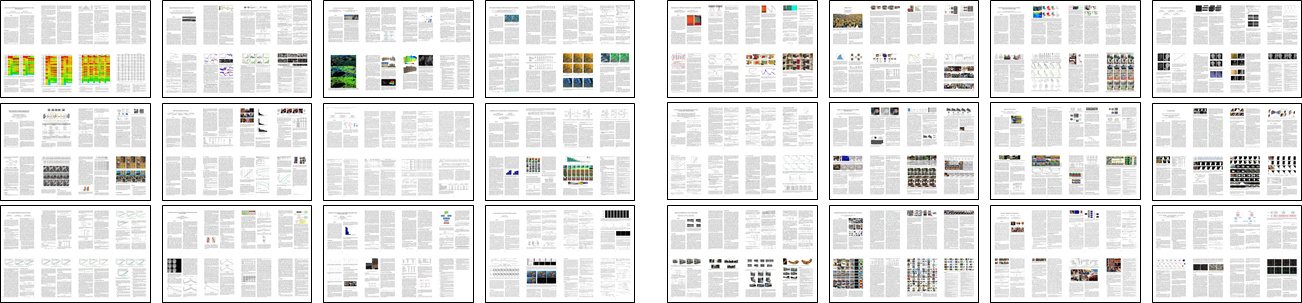} 
\mpage{0.49}{Workshop papers} \hfill
\mpage{0.49}{Conference papers}
\vspace{\figcapmargin}
\caption{\tb{Random samples of the collected Computer Vision Paper Gestalt (CVPG) datasets}. Glancing through samples in both classes show that there are differences in terms of the general layout of the paper. Our goal here is to leverage deep ConvNets to learn representation for capturing these patterns.}
\label{fig:dataset}
\end{figure*}

The detailed statistics of the collected Computer Vision Paper Gestalt (CVPG) dataset are shown in \tabref{stat}. 
There are in total 5618 positive examples and 1503 negative examples.
\figref{dataset} shows random samples from both positive and negative samples of the collected dataset.
The dataset is available on our project website~\url{https://github.com/vt-vl-lab/paper-gestalt}.

\begin{table}[t] \centering
\caption{\tb{Computer Vision Paper Gestalt (CVPG) dataset.} }
\label{tab:stat}
\resizebox{\linewidth}{!}{
\begin{tabular}{lclc}
\toprule
\multicolumn{2}{c}{Positive examples} &
\multicolumn{2}{c}{Negative examples} \\
Venue & \# samples & Venue & \# samples\\
\midrule 
CVPR 2013 & 471 & CVPR-W 2013 & 80 \\
ICCV 2013 & 454 & ICCV-W 2013 & 101\\
CVPR 2014 & 540 & CVPR-W 2014 & 61\\
CVPR 2015 & 602 & CVPR-W 2015 & 113\\
ICCV 2015 & 526& ICCV-W 2015 & 116\\
CVPR 2016 & 643& CVPR-W 2016 & 184\\
ICCV 2017 & 621& ICCV-W 2017 & 350\\
CVPR 2017 & 783& CVPR-W 2017 & 251\\
CVPR 2018 & 978& CVPR-W 2018 & 247\\
\midrule
Total & 5618 & Total & 1503 \\
\bottomrule
\end{tabular}
}
\end{table}

\subsection{Paper review as image classification}
To simulate the actual potential usage of our system (i.e., predicting good/bad papers from unseen paper submissions), we use the positive/negative examples in the CVPR 2018 as our testing set and the papers in the prior conferences/workshops from 2013 to 2017 as our training set.

We use ResNet-18~\cite{he2016deep} (pre-trained on ImageNet) as our classification network.\footnote{Deeper networks with larger capacity can also be used. However, we do not observe significant performance improvement when using ResNet-34 or ResNet-50.}
We replace the ImageNet 1,000 class classification head with two output nodes (good or bad papers).
Following the practice of transfer learning, we finetune the ImageNet pre-trained network on the proposed CVPG dataset with stochastic gradient descent (SGD) with a momentum of 0.9 for a total of 50 epochs.
We set the initial learning rate as 0.001 and decay it by a factor of 0.1 every ten epochs.
To accommodate the class-imbalanced training data, we use the weighted cross-entropy loss (weighted by the inverse of the training examples in each class).
We resize all the images to $224 \times 224$ pixels for both training and testing.
We choose not to apply standard data augmentation techniques such as random cropping, horizontal flipping, or photometric transformation during training to keep the original visual content and layout of the entire paper.
The network training process takes less than 30 minutes on a NVIDIA Titan V100 GPU.

\subsection{Experimental results}

\begin{figure}[t]
\includegraphics[width =\linewidth]{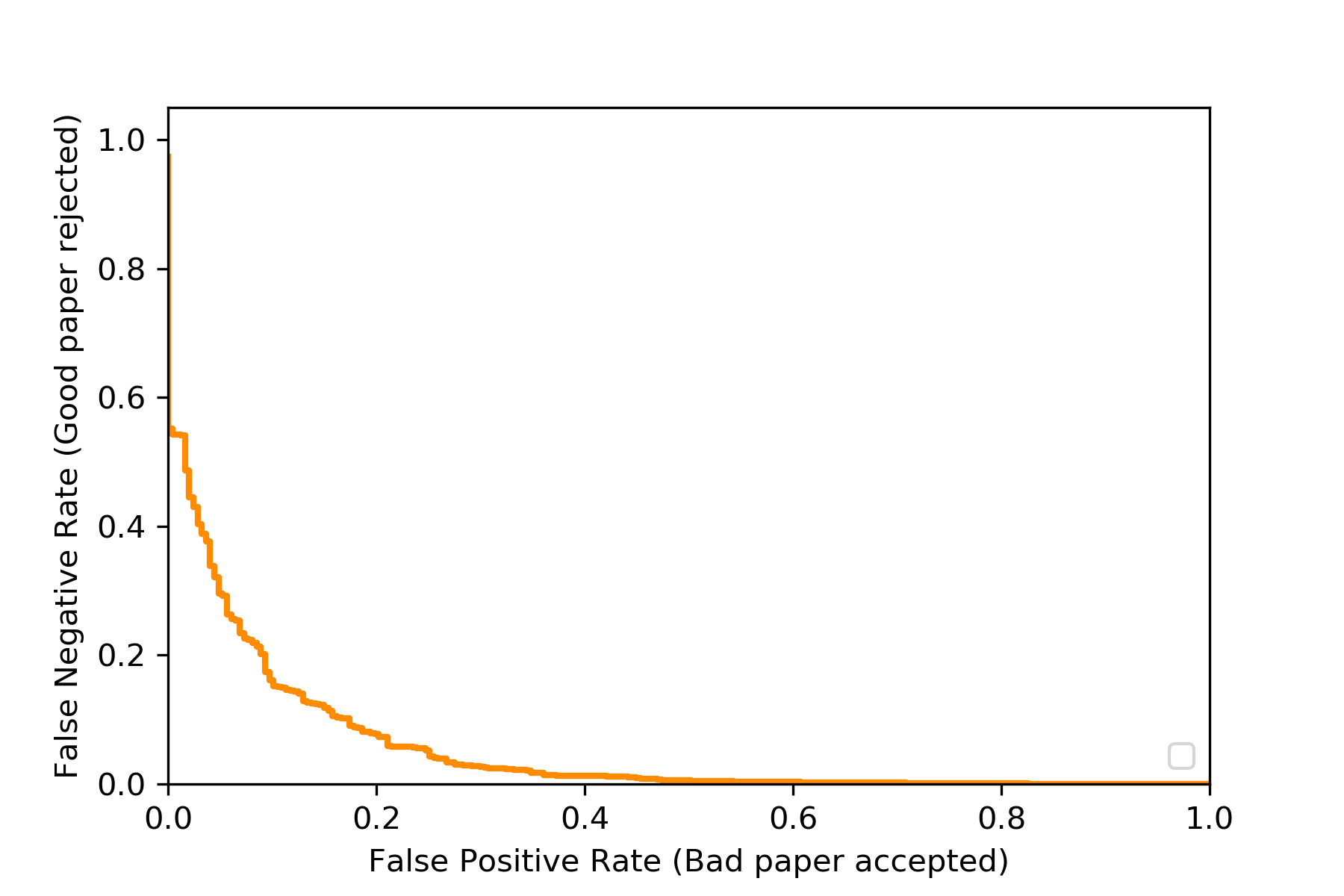}
\vspace{\figcapmargin}
\caption{\textbf{Performance characterization of the trained good/bad paper classifier.} The x-axis denotes the false positive rate (the percentage of bad papers getting accepted). The y-axis denotes the false negative rate (the percentage of good papers getting rejected).}
\label{fig:roc}
\end{figure}

\paragraph{Evaluation.} On the test dataset (CVPR 2018 conference/workshop papers), our trained network achieves an overall accuracy of 92\%.
By varying the threshold on the network predictions after the softmax layer, we plot the ROC curve to further characterize the performance of our model in \figref{roc}.
The x-axis shows the \emph{false positive rate (FPR)}, indicating the portion of bad papers getting accepted by our model.
The y-axis shows the \emph{false negative rate (FNR)}, indicating the portion of good papers getting rejected by our model.
Note that here we plot the FNR instead of the true positive rate (TPR) to better illustrate the trade-off between the two error types.

\begin{figure*}[t]
\mfigure{0.235}{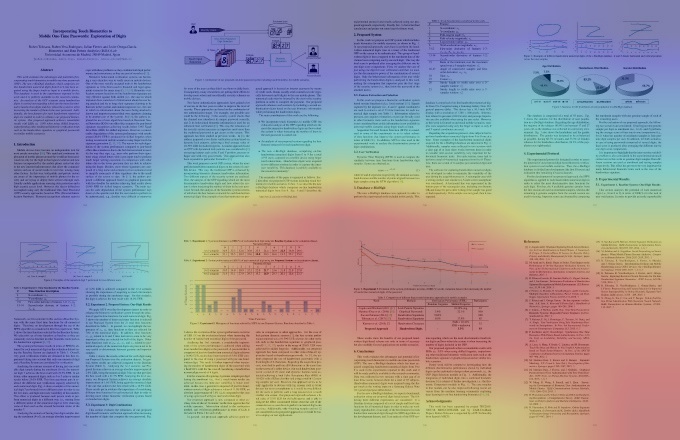} \hfill
\mfigure{0.235}{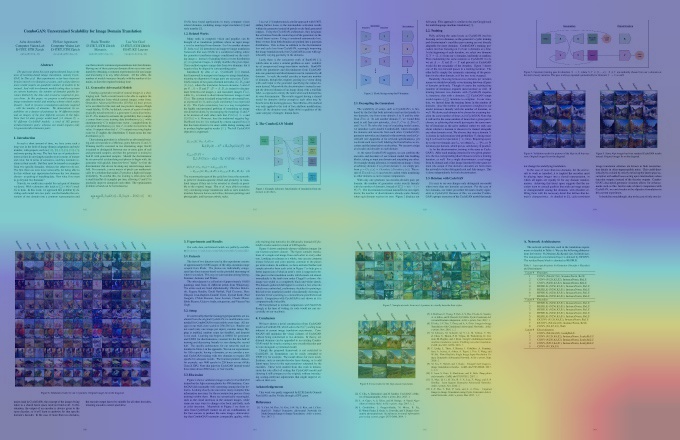} \hfill
\mfigure{0.235}{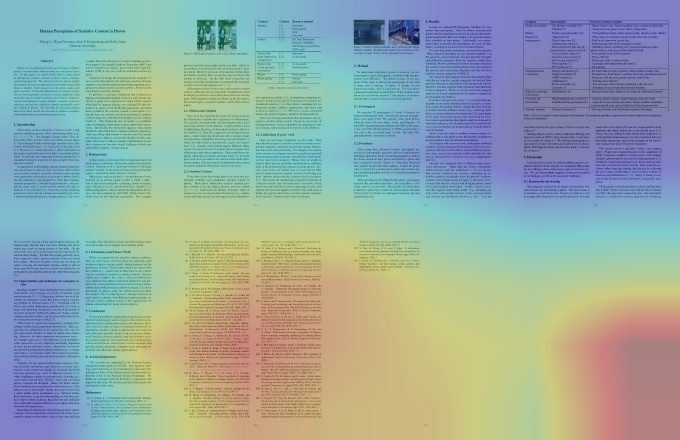} \hfill
\mfigure{0.235}{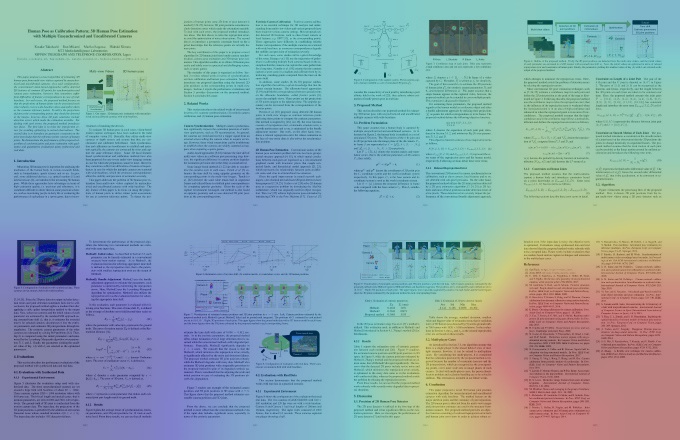} \\
\vspace{0.5mm} \\
\mfigure{0.235}{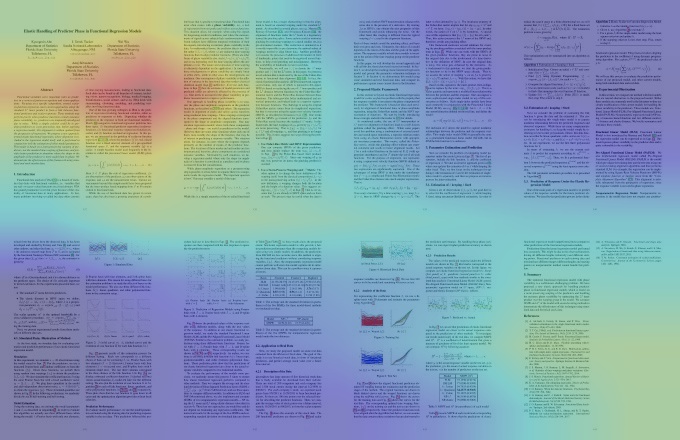} \hfill
\mfigure{0.235}{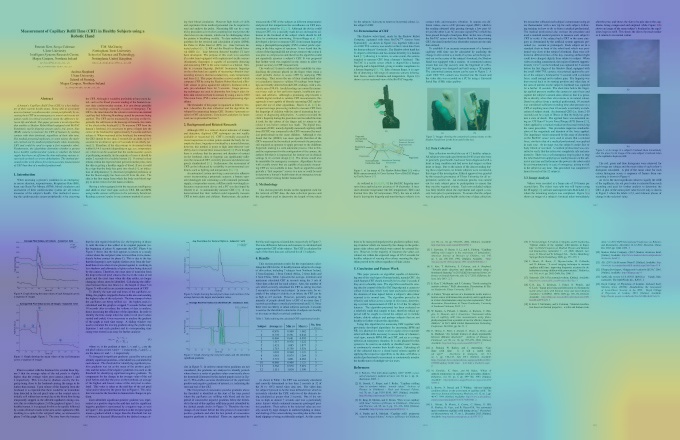} \hfill
\mfigure{0.235}{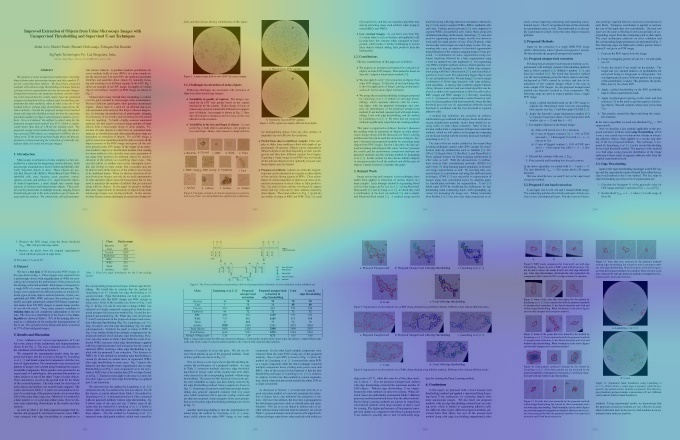} \hfill
\mfigure{0.235}{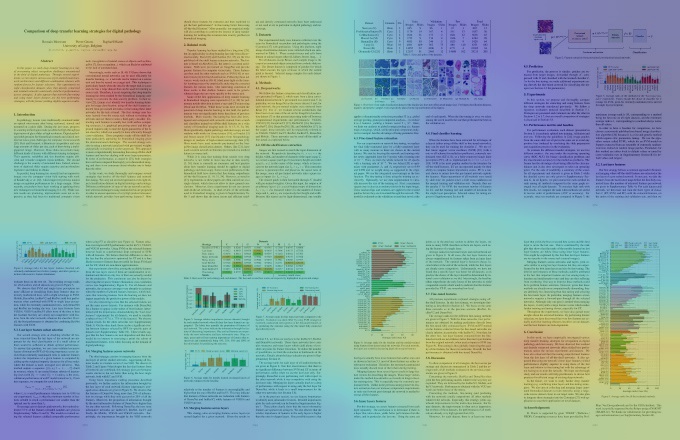} \\
\vspace{\figcapmargin}
\caption{\tb{Class-specific discriminative regions for \emph{bad} papers.} 
(\emph{Top}) Failing to fill the paper into a full eight-page paper is a discriminative visual cue for bad paper.
(\emph{Bottom}) The generated heatmaps focus on the top-right corner of the first page. This suggests that the \emph{absence} of illustrative figures in the first two pages may cause the paper more difficult to understand.
}
\label{fig:CAM_workshop}
\end{figure*}
\begin{figure*}[t]
\mfigure{0.235}{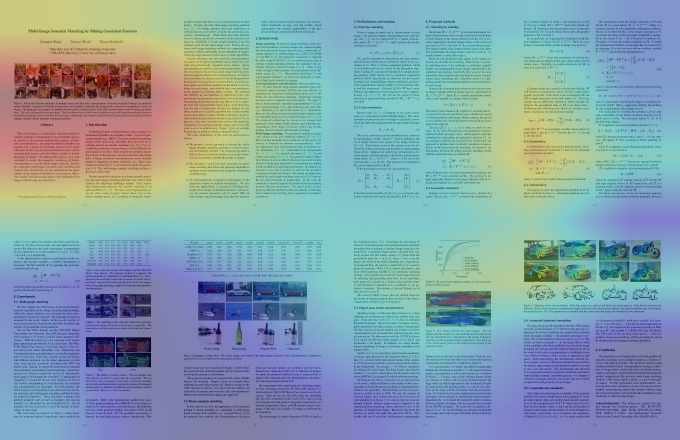} \hfill
\mfigure{0.235}{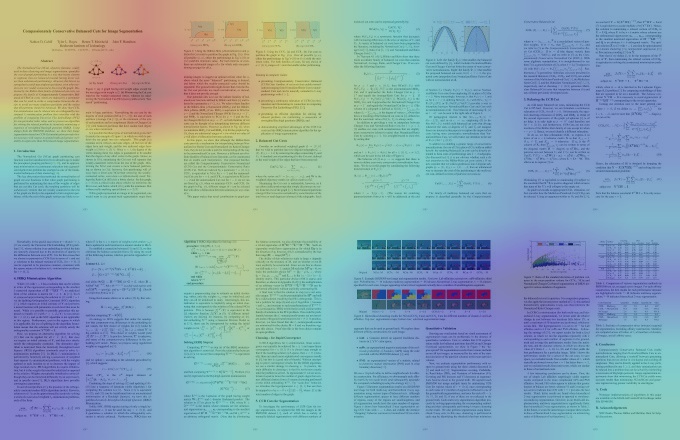} \hfill
\mfigure{0.235}{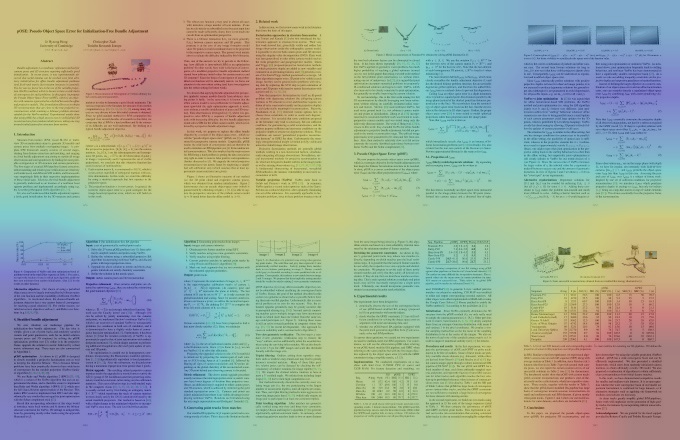} \hfill
\mfigure{0.235}{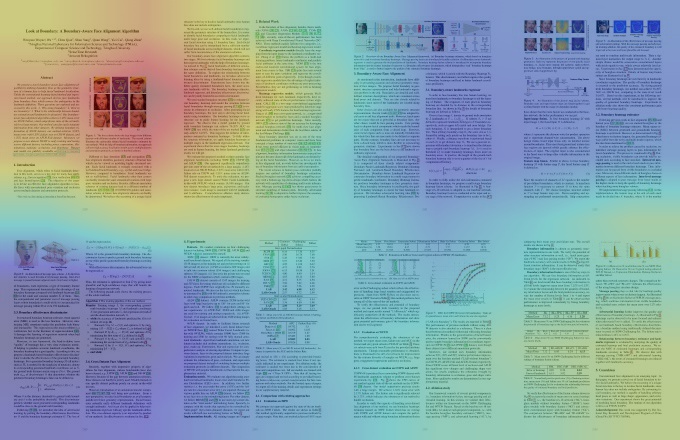} \\
\vspace{0.5mm} \\
\mfigure{0.235}{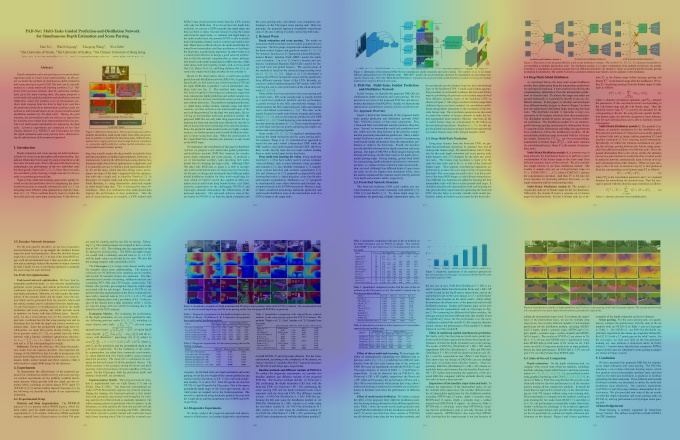} \hfill
\mfigure{0.235}{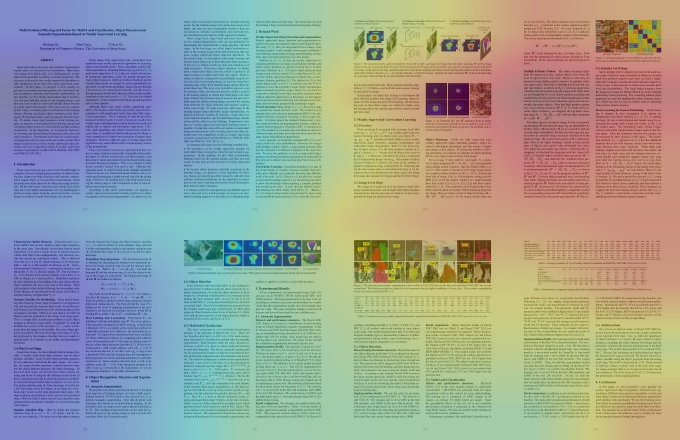} \hfill
\mfigure{0.235}{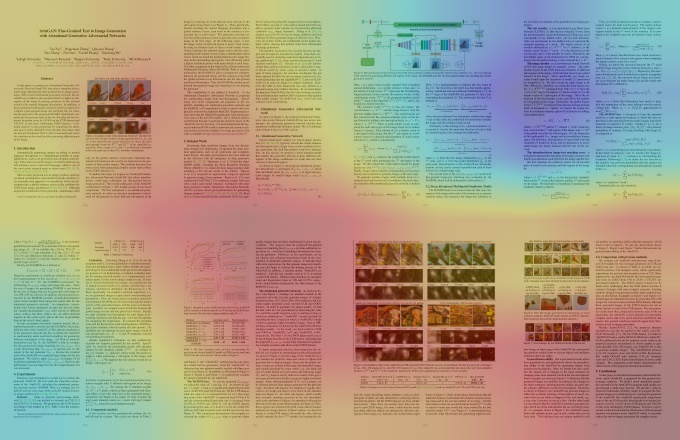} \hfill
\mfigure{0.235}{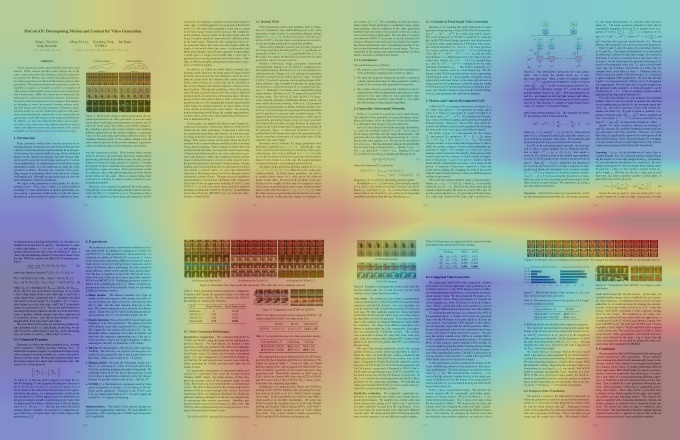} \\
\vspace{\figcapmargin}
\caption{\tb{Class-specific discriminative regions for \emph{good} papers.} The heatmap generated by class activation mapping~\cite{zhou2016learning} highlights regions specific to good papers, e.g., teaser figures in the first page for illustrating the main ideas, tables/plots showing a sense of thoroughness in experimental validation, impressive math equations, and arrays of colorful images for qualitative results from benchmark datasets.
}
\label{fig:CAM_conference}
\end{figure*}

Choosing different threshold values leads to different trade-off.
For example, if we allow only 0.4\% of the good papers getting rejected, we can accurately reject 50\% of the bad ones.
If we allow 5\% of the good papers getting rejected (as there will be inevitable noises in peer reviews anyway), we can reject up to 75\% of the bad papers.

Here we use a more concrete example to better understand what the results mean.
There are in total 3309 valid submissions to CVPR 2018 with 979 of them accepted (good papers) and 2230 papers rejected (bad papers).
Assuming the actual negative examples show the same distributions in the workshop papers, applying our model (with 0.4\% FPR and 50\% FNR) to all the valid submissions to CVPR 2018 can safely reject 1115 bad papers (without peer reviews) at the cost of sacrificing 4 good papers (among the 979 good ones).
Such an automatic pre-filtering stage substantially reduces the workload of reviewers.







\paragraph{Class-specific activation maps.} 
While the model achieves decent classification performance, we believe that it is unlikely that the classifier will ever be used in an actual conference.
Nevertheless, we can take a closer look at how the classification model makes the decision and in turns improve the paper gestalt of our future paper submissions.

Multiple visualization techniques for understanding deep neural networks have been proposed.
For examples, retrieving image patches that maximize a particular neuron~\cite{girshick2014rich}, reconstructing the input image~\cite{mahendran2015understanding}, quantifying the interpretability of latent layers~\cite{bau2017network}, and mapping the activations to the input image space with a deconvNet~\cite{zeiler2014visualizing}.
In this paper, we use the class-specific activation mapping~\cite{zhou2016learning} for visualizing the discriminative regions for the classifying a paper into a good or a bad one.

\figref{CAM_workshop} shows sample class-specific activation maps on papers that were rejected by our model.
In the first row, the discriminative regions generated by our classifier highlight mostly on the \emph{incomplete pages}.
This makes sense because well-polished papers often squeeze the contents compactly into precisely 8 pages (with some clever use of \texttt{vspace}).
In the second row, the class-specific activation maps focus on the top-right corner of the first page.
It appears that the classifier picks up the \emph{absence of a motivation or teaser figure} in the first page as its primary reason for rejecting a paper.
This reveals that it is crucial to include a motivation figure on the first page to illustrate the main idea of the work.

\figref{CAM_conference}, on the other hand, shows the class-specific activation maps on papers that were accepted by our model.
The discriminative regions for good papers include the teaser figure (easier to understand the main idea), detailed tables (comprehensive experimental evaluation), and colorful images (qualitative results).
We believe that such visualization can be applied as a diagnostic tool to help identify the strength/weakness of one's paper submissions in the future.

\begin{figure}[t] \centering
\includegraphics[width =\linewidth]{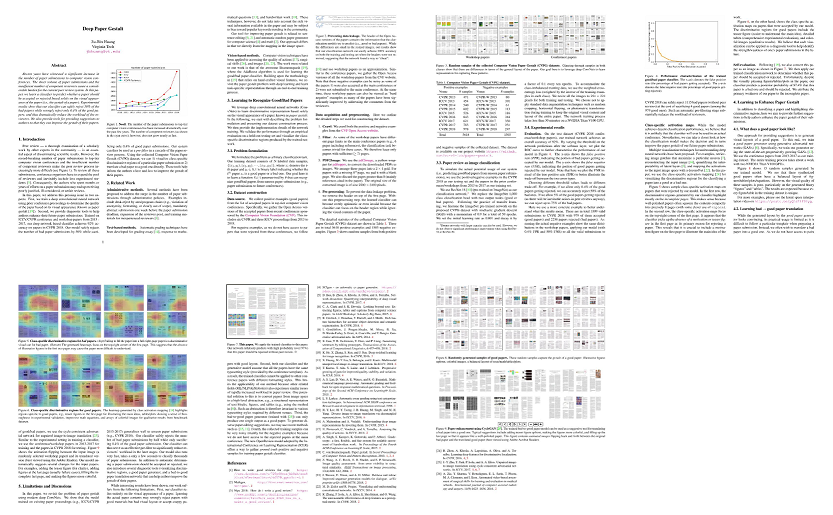} 
\vspace{\figcapmargin}
\caption{\tb{This paper.} We apply the trained classifier to this paper. Our network ruthlessly predicts with high probability (over 97\%) that this paper should be rejected without peer review. \frownie
}
\label{fig:self}
\end{figure}

\paragraph{Self-evaluation.} Following \cite{Bearnensquash2010}, we also convert this paper as an image as shown in \figref{self}.
We then apply our trained classification network to determine whether this paper should be accepted or rejected. 
Unfortunately, despite the visually pleasing figures/tables/plots in the paper, our classifier predicts a posterior probability of 97.4\% that this paper is a bad one and should be rejected. 
We attribute the primary weakness of our paper to the incomplete pages.

\section{Learning to Enhance Paper Gestalt}

In addition to classifying a paper and highlighting discriminative regions, here we aim to provide further suggestions to help authors enhance the paper gestalt of their submissions.
\begin{figure*}[t] \centering
\includegraphics[width =\linewidth]{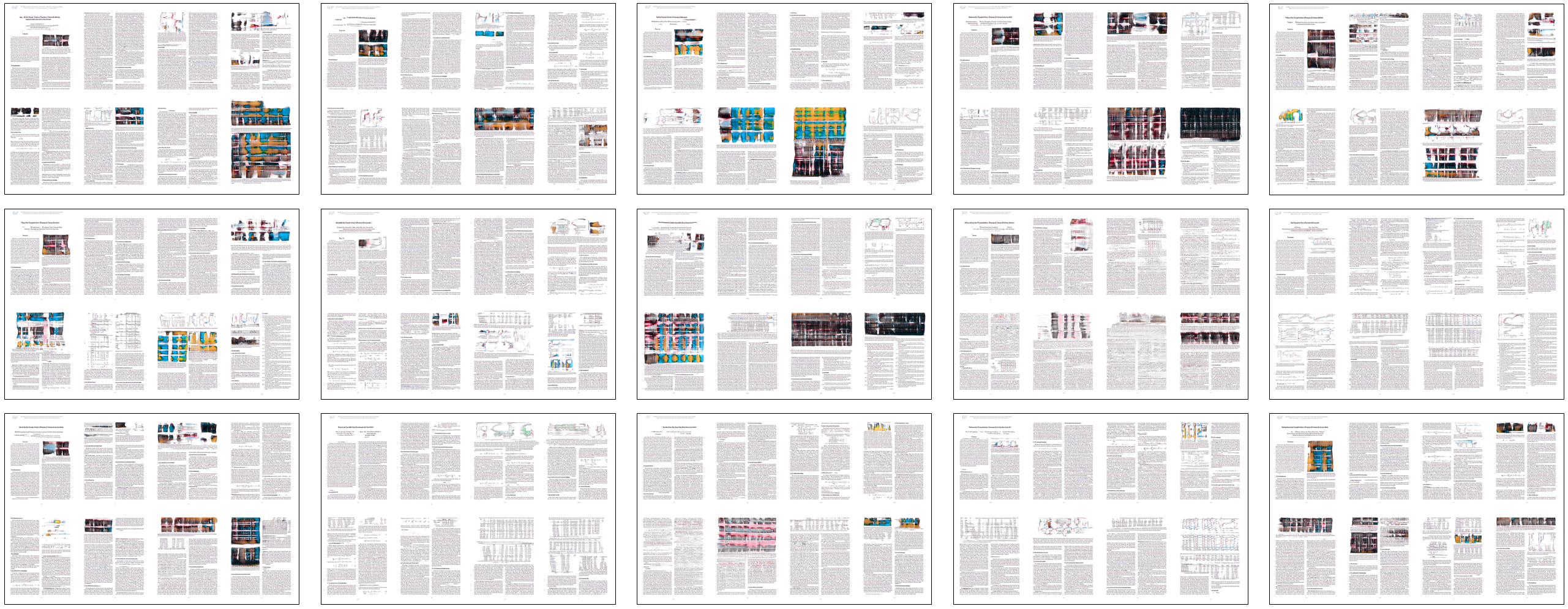} 
\vspace{\figcapmargin}
\caption{\tb{Randomly generated samples of good papers.} These random  samples capture the \emph{gestalt} of a good paper: illustrative figures upfront, colorful images, a balanced layout of texts/math/tables/plots.}
\label{fig:prog_gan}
\end{figure*}

\subsection{What does a good paper look like?}
One approach for providing suggestions is to generate visual layouts of a good paper.
To this end, we train a \emph{good paper generator} using generative adversarial networks (GANs)~\cite{goodfellow2014generative}.
Specifically, we train our generator using the state-of-the-art progressively growing GANs~\cite{karras2018progressive}.
We use the conference papers from 2013-2017 as our training dataset.
The entire training process takes about a week with two NVIDIA Titan V100 GPUs.

\figref{prog_gan} shows 15 random samples generated by our trained model.
We see that these synthesized good papers often have a balanced layout of figures/tables/plots/equations.
However, the visual quality of these samples is poor, particularly on the generated blurry ``figures'' and ``tables''.
The results are expected because every figure/table in the training dataset is unique.

For more examples, please see the latent space interpolation video on \url{https://youtu.be/yQLsZLf02yg}.

\begin{figure*}[t]
\centering
\begin{animateinline}[autoplay,loop,controls]{1}
  \centering
  \multiframe{2}{i=1+1}{%
      \includegraphics[width=0.32\linewidth]{figs/cyclegan/02/000\i.png} \hfill
      \includegraphics[width=0.32\linewidth]{figs/cyclegan/04/000\i.png} \hfill
      \includegraphics[width=0.32\linewidth]{figs/cyclegan/03/000\i.png}%
  }
\end{animateinline}
\vspace{\figcapmargin}
\caption{\textbf{Paper enhancement using CycleGAN~\cite{zhu2017unpaired}.} The trained bad-to-good paper model can be used as a suggestive tool for translating a bad paper into a good one. Typical suggestions include adding teaser figure upfront, making the figures more colorful, and filling up the last page so that it appears like a well-polished paper.
This figure contains \emph{animated images} flipping back and forth between the original bad paper and the translated good paper (best viewed using Adobe Acrobat Reader).
}
\label{fig:cycleGAN}
\end{figure*}

\subsection{Learning bad $\rightarrow$ good paper translation}

While the generated layout by the \emph{good paper generator} looks convincing, its practical usage is limited as it is difficult to follow a particular template when preparing a paper submission.
Instead, we often wish to \emph{translate} a bad paper into a good one.
As we do not have access to pairs of good/bad papers, we use the cycle-consistent adversarial network for unpaired image-to-image translation~\cite{zhu2017unpaired}.
Similar to the experimental setting in training a classifier, we use the conference/workshop papers in 2013-2017 for training and the papers in CVPR 2018 for testing.
\figref{cycleGAN} shows the animation flipping between the input image (a randomly selected workshop paper) and its translated version (best viewed using the Adobe Reader).
Our model automatically suggests several changes for the input papers.
For examples, adding the teaser figure (for clarity), adding figures at the last page (usually failure cases), filling the incomplete last page, and making the figures more colorful.

\section{Limitations and Discussions}

In this paper, we revisit the problem of paper gestalt using modern deep ConvNets. 
We show that the model trained on existing paper proceedings (e.g., ICCV/CVPR 2013-2017) generalizes well to unseen paper submissions (e.g., CVPR 2018).
Our classifier safely rejects the number of bad paper submissions by half while only sacrificing 0.4\% of the good paper submission.
Our classifier can thus serve as an effective pre-filter to significantly reduce reviewers' workload in the later stages.
Our model also runs very fast, takes a only a few seconds to classify thousands of paper submissions.
In addition to automatic determining a paper submission should be accepted or rejected, we also introduce several diagnostic tools (visualizing discriminative regions, a good paper generator, and a bad-to-good paper translation network) that can help authors improve the gestalt of their papers.

While interesting results have been shown, our work suffers from the following limitations.
First, our classifier relies entirely on the visual appearance of a paper.
Ignoring the actual paper contents may wrongly reject papers with good materials but bad visual layout or accept crappy papers with good layout.
Second, both our classifier and the generative model assume that all the papers have the same typesetting style (provided by the conference template).
As a result, the trained classifier cannot be applied to other conference papers with different formatting styles.
This limits the applicability of our method because other related fields (ML/NLP/AI/Robotics) also experience similar issues of rapidly increased workload in paper review.
One potential solution to this is to convert papers from image space to a high-level abstraction, e.g., a structured representation of text blocks, figures, and tables (e.g., using the method in~\cite{clark2015looking}).
Such an abstraction is therefore invariant to various typesetting styles required by different venues.
Third, the bad-to-good paper generator (trained with \cite{zhu2017unpaired}) can only produce one single output as a good paper.
To generate diverse paper editing suggestions, we may use recent methods such as ~\cite{lee2018diverse,huang2018multimodal}.
Fourth, the collected training samples can be very noisy (mainly for the negative examples) because we do not have access to the rejected papers at the main conference. 
The new OpenReview model adopted by the International Conference on Learning Representation (ICLR) offers a way to gather \emph{ground truth} positive and negative samples for training paper gestalt classifier.

{\small
\bibliographystyle{ieee}
\bibliography{main}

\begin{thebibliography}{10}\itemsep=-1pt

\bibitem{CVPRreview}
How to write good reviews for cvpr.
\newblock
  \url{https://www.dropbox.com/s/725p60wcajbb8xh/How\%20to\%20Review\%20for\%20CVPR.pptx?dl=0}.

\bibitem{Mathgen}
Mathgen.
\newblock \url{http://thatsmathematics.com/mathgen/}.

\bibitem{NIPS2018}
Nips 2018: How do i write a good review?
\newblock
  \url{https://www.reddit.com/r/MachineLearning/comments/8ite3n/r_nips_2018_how_do_i_write_a_good_review/}.

\bibitem{SCIgen}
{SCIgen} - an automatic cs paper generator.
\newblock \url{https://pdos.csail.mit.edu/archive/scigen/}.

\bibitem{bau2017network}
D.~Bau, B.~Zhou, A.~Khosla, A.~Oliva, and A.~Torralba.
\newblock Network dissection: Quantifying interpretability of deep visual
  representations.
\newblock In {\em CVPR}, 2017.

\bibitem{clark2015looking}
C.~A. Clark and S.~K. Divvala.
\newblock Looking beyond text: Extracting figures, tables and captions from
  computer science papers.
\newblock In {\em AAAI Workshop: Scholarly Big Data}, 2015.

\bibitem{girshick2014rich}
R.~Girshick, J.~Donahue, T.~Darrell, and J.~Malik.
\newblock Rich feature hierarchies for accurate object detection and semantic
  segmentation.
\newblock In {\em CVPR}, 2014.

\bibitem{goodfellow2014generative}
I.~Goodfellow, J.~Pouget-Abadie, M.~Mirza, B.~Xu, D.~Warde-Farley, S.~Ozair,
  A.~Courville, and Y.~Bengio.
\newblock Generative adversarial nets.
\newblock In {\em NIPS}, 2014.

\bibitem{guu2018generating}
K.~Guu, T.~B. Hashimoto, Y.~Oren, and P.~Liang.
\newblock Generating sentences by editing prototypes.
\newblock {\em Transactions of the Association of Computational Linguistics},
  6:437--450, 2018.

\bibitem{he2016deep}
K.~He, X.~Zhang, S.~Ren, and J.~Sun.
\newblock Deep residual learning for image recognition.
\newblock In {\em CVPR}, 2016.

\bibitem{huang2018multimodal}
X.~Huang, M.-Y. Liu, S.~Belongie, and J.~Kautz.
\newblock Multimodal unsupervised image-to-image translation.
\newblock In {\em ECCV}, 2018.

\bibitem{karras2018progressive}
T.~Karras, T.~Aila, S.~Laine, and J.~Lehtinen.
\newblock Progressive growing of gans for improved quality, stability, and
  variation.
\newblock In {\em ICLR}, 2018.

\bibitem{lan2015mathematical}
A.~S. Lan, D.~Vats, A.~E. Waters, and R.~G. Baraniuk.
\newblock Mathematical language processing: Automatic grading and feedback for
  open response mathematical questions.
\newblock In {\em Proceedings of the Second ACM Conference on Learning@ Scale},
  2015.

\bibitem{larkey1998automatic}
L.~S. Larkey.
\newblock Automatic essay grading using text categorization techniques.
\newblock In {\em International ACM SIGIR conference on Research and
  development in information retrieval}, 1998.

\bibitem{lee2018diverse}
H.-Y. Lee, H.-Y. Tseng, J.-B. Huang, M.~Singh, and M.-H. Yang.
\newblock Diverse image-to-image translation via disentangled representations.
\newblock In {\em ECCV}, 2018.

\bibitem{mahendran2015understanding}
A.~Mahendran and A.~Vedaldi.
\newblock Understanding deep image representations by inverting them.
\newblock In {\em CVPR}, 2015.

\bibitem{pirsiavash2014assessing}
H.~Pirsiavash, C.~Vondrick, and A.~Torralba.
\newblock Assessing the quality of actions.
\newblock In {\em ECCV}, 2014.

\bibitem{singh2017gradescope}
A.~Singh, S.~Karayev, K.~Gutowski, and P.~Abbeel.
\newblock Gradescope: a fast, flexible, and fair system for scalable assessment
  of handwritten work.
\newblock In {\em Proceedings of the Fourth ACM Conference on Learning@ Scale},
  2017.

\bibitem{Bearnensquash2010}
C.~von Bearnensquash.
\newblock Paper gestalt.
\newblock In {\em Secret Proceedings of Computer Vision and Pattern
  Recognition}, 2010.

\bibitem{wang2004image}
Z.~Wang, A.~C. Bovik, H.~R. Sheikh, and E.~P. Simoncelli.
\newblock Image quality assessment: from error visibility to structural
  similarity.
\newblock {\em IEEE Transactions on image processing}, 13(4):600--612, 2004.

\bibitem{weston2018retrieve}
J.~Weston, E.~Dinan, and A.~H. Miller.
\newblock Retrieve and refine: Improved sequence generation models for
  dialogue.
\newblock {\em arXiv preprint arXiv:1808.04776}, 2018.

\bibitem{zeiler2014visualizing}
M.~D. Zeiler and R.~Fergus.
\newblock Visualizing and understanding convolutional networks.
\newblock In {\em ECCV}, 2014.

\bibitem{zhang2018unreasonable}
R.~Zhang, P.~Isola, A.~A. Efros, E.~Shechtman, and O.~Wang.
\newblock The unreasonable effectiveness of deep features as a perceptual
  metric.
\newblock In {\em CVPR}, 2018.

\bibitem{zhou2016learning}
B.~Zhou, A.~Khosla, A.~Lapedriza, A.~Oliva, and A.~Torralba.
\newblock Learning deep features for discriminative localization.
\newblock In {\em CVPR}, 2016.

\bibitem{zhu2017unpaired}
J.-Y. Zhu, T.~Park, P.~Isola, and A.~A. Efros.
\newblock Unpaired image-to-image translation using cycle-consistent
  adversarial networks.
\newblock In {\em ICCV}, 2017.

\bibitem{zia2016automated}
A.~Zia, Y.~Sharma, V.~Bettadapura, E.~L. Sarin, T.~Ploetz, M.~A. Clements, and
  I.~Essa.
\newblock Automated video-based assessment of surgical skills for training and
  evaluation in medical schools.
\newblock {\em International journal of computer assisted radiology and
  surgery}, 11(9):1623--1636, 2016.

\end{thebibliography}
}

\end{document}